# An Approximate Nonmyopic Computation for Value of Information[*]

**David Heckerman**
Computer Science and Pathology
University of Southern California
HMR 204, 2025 Zonal Ave
Los Angeles, CA 90033

**Eric Horvitz**
Palo Alto Laboratory
Rockwell Int Science Center
444 High Street
Palo Alto, California 94301

**Blackford Middleton**
Section on Medical Informatics
Division of General Internal Medicine
Stanford University Medical Center
Stanford, California 94305

## Abstract

Value-of-information analyses provide a straightforward means for selecting the best next observation to make, and for determining whether it is better to gather additional information or to act immediately. Determining the next best test to perform, given a state of uncertainty about the world, requires a consideration of the value of making all possible sequences of observations. In practice, decision analysts and expert-system designers have avoided the intractability of exact computation of the value of information by relying on a *myopic* approximation. Myopic analyses are based on the assumption that only one additional test will be performed, even when there is an opportunity to make a large number of observations. We present a nonmyopic approximation for value of information that bypasses the traditional myopic analyses by exploiting the statistical properties of large samples.

## 1 INTRODUCTION

A person faced with a decision usually has the opportunity to gather additional information about the state of the world before taking action. Decision-theoretic methods for determining the value of gathering additional information date back to the earliest literature on the principle of maximum expected utility (MEU). These methods form an integral part of many probabilistic expert systems, such as Gorry's congestive-heart-failure program (Gorry and Barnett, 1968) and Pathfinder (Heckerman et al., 1989; Heckerman et al., 1990), an expert system that assists pathologists with the diagnosis of lymph-node diseases. To decide whether or not to perform a test, an expert system computes the value of information of that test. The system recommends that the test be performed if and only if the value of information exceeds the cost of the test.[1]

In most decision contexts, a decision maker has the option to perform several tests, and can decide which test to perform after seeing the results of all previous tests. Thus, an expert system should consider the value of all possible *sequences* of tests. Such an analysis is intractable, because the number of sequences grows exponentially with the number of tests. Builders of expert systems have avoided the intractability of complete value-of-information analyses by implementing *myopic* or *greedy* value-of-information analyses. In such analyses, a system determines the next best test by computing value of information based on the assumption that the decision maker will act immediately after seeing the results of the single test (Gorry et al., 1973; Heckerman et al., 1990). In this paper, we present an approximate nonmyopic analysis. The analysis avoids the traditional myopic assumption by making use of the statistical properties of large samples.

## 2 VALUE-OF-INFORMATION COMPUTATIONS FOR DIAGNOSIS

We discuss myopic and nonmyopic value-of-information computations in terms of the simple model for diagnosis under uncertainty represented by the influence diagram in Figure 1. In this model, the chance node $H$ represents a mutually exclusive and exhaustive set of possible hypotheses, and the decision node $D$ represents a mutually exclusive and exhaustive set of possible alternatives. The value node $U$ represents the utility of the decision maker, which depends on the outcome of $H$ and the decision $D$. The chance nodes $E_1$, $E_2$, ..., $E_n$ are observable pieces of evi-dence or tests about the true state of $H$. This model is identical to that for Pathfinder (Heckerman, 1990).

---

[1]This prescription for action assumes that the delta property holds. See Section 3.

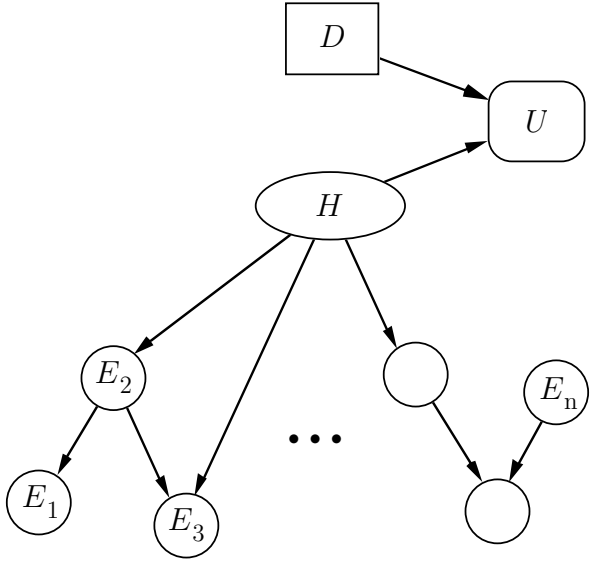

Figure 1: An influence-diagram representation of the problem of diagnosis under uncertainty. The decision-maker's utility (rounded rectangular node, $U$) depends on a hypothesis (oval node, $H$) and a decision (square node, $D$). The variables $E_i$ are pieces of evidence or tests about the true state of $H$.

We make several simplifying assumptions. First, we assume that $H$ is a binary chance variable and $D$ is a binary decision variable. We use $H$ and $\neg H$ to denote the two outcomes of $H$, and $D$ and $\neg D$ to denote the two outcomes of $D$. For definiteness, we assume that the decision maker chooses $D$ (as opposed to $\neg D$), when $H$ occurs. Second, we assume that each piece of evidence, $E_1, E_2, \ldots, E_n$, is binary. Finally, we assume that each piece of evidence is conditionally independent of all other evidence, given $H$ and $\neg H$. In Section 6, we relax these assumptions.

Using the assumption of conditional independence of evidence, we can calculate the posterior probability of the hypothesis by multiplying together all of the likelihood ratios, $\frac{p(E_i|H)}{p(E_i|\neg H)}$, with the prior odds, $\frac{p(H)}{p(\neg H)}$.

$$\frac{p(H|E_i,\ldots,E_m)}{P(\neg H|E_i,\ldots,E_m)} = \frac{p(E_1|H)}{p(E_1|\neg H)} \cdots \frac{p(E_m|H)}{p(E_m|\neg H)} \frac{p(H)}{p(\neg H)}$$

We can write this equation more compactly in odds form as

$$O(H|E_i,\ldots,E_m) = O(H) \prod_{i=1}^{m} \lambda_i \quad (1)$$

where $\lambda_i$ is the likelihood ratio $\frac{p(E_i|H)}{p(E_i|\neg H)}$.

Because $D$ and $H$ are binary, it follows from the MEU principle that there exists a threshold probability $p^*$, such that we should take action $D$ if and only if the probability of $H$ exceeds $p^*$. This threshold is the probability of $H$ at which the decision maker is indifferent between acting and not acting. That is, $p^*$ is the point where acting and not acting have equal utility, or

$$\begin{aligned} p^*U(H,D) + (1-p^*)U(\neg H, D) = \\ p^*U(H,\neg D) + (1-p^*)U(\neg H, \neg D) \end{aligned} \quad (2)$$

In Equation 2, $U(H, D)$ is the decision maker's utility for the situation where $H$ occurs and action $D$ is taken, $U(H, \neg D)$ is the utility when $H$ occurs and action $D$ is not taken, and so on. Solving Equation 2 for $p^*$, we obtain

$$p^* = \frac{C}{C+B} \quad (3)$$

where $C$ is the *cost* of the decision

$$C = U(\neg H, \neg D) - U(\neg H, D) \quad (4)$$

and $B$ is the *benefit* of the decision

$$B = U(H, D) - U(H, \neg D) \quad (5)$$

If the decision maker has observed pieces of evidence $E_1, E_2, \ldots, E_m$, then the decision maker should choose action $D$ if and only if $p(H|E_1 \ldots, E_m) > p^*$. In terms of the odds formulation, he should act if and only if

$$O(H|E_1,\ldots,E_m) \geq \frac{p^*}{1-p^*} \quad (6)$$

The weight of evidence, $w_i$, is defined as the log of the likelihood ratio, $\ln \lambda_i$. Mapping likelihood ratios into weights of evidence allows us to update the probability of $H$ through the addition of the weights of evidence. Referring to Equations 1 and 6, we can rewrite the threshold-probability condition in terms of the log-likelihood ratio where $w_i = \ln \lambda_i$. The decision maker should choose action $D$ if and only if

$$W = \sum_{i=1}^{m} w_i \geq \ln \frac{p^*}{1-p^*} - \ln O(H) = W^* \quad (7)$$

In this expression, $W^*$ is the decision threshold in terms of weights of evidence.

## 3 MYOPIC ANALYSIS

Let us assume that the user of a diagnostic system has instantiated zero or more pieces of evidence in the influence diagram shown in Figure 1. We can propagate the effects of these instantiations to the uninstantiated nodes, and remove the instantiated nodes from the influence diagram. This removal leaves an influence diagram of the same form as that shown in Figure 1. To simplify our notation, we continue to refer to the remaining pieces of evidence as $E_1, E_2, \ldots, E_n$; also, we use $p(H)$ to refer to the probability of the hypothesis $H$, given the instantiated evidence.

The decision maker now considers whether he should observe another piece of evidence before acting. A myopic procedure for identifying such evidence computes, for each piece of evidence, the expected utility of the decision maker under the assumption that *the decision maker will act after observing only that piece of evidence.* In addition, the procedure computes his expected utility if he does not observe any evidence before making his decision. If, for each piece of evidence, the expected utility given that evidence is less than the expected utility given no evidence, then the decision maker acts immediately in accordance with Equation 6. Otherwise, the decision maker observes the piece of evidence with the highest expected utility; then, the myopic procedure *repeats this computation to identify additional evidence for observation.* Because the myopic procedure allows for the gathering of additional evidence, the procedure is inconsistent with its own assumptions. We return to this observation in the next section.

In the remainder of this section, we examine the computation of expected utilities and introduce notation. Let $EU(E, C_E)$ denote the expected utility of the decision maker who will observe $E$ at cost $C_E$, and then act. Let $CE(E, C_E)$ be the certain equivalent of this situation. That is,

$$U(CE(E, C_E)) = EU(E, C_E) \qquad (8)$$

or

$$CE(E, C_E) = U^{-1}(EU(E, C_E)) \qquad (9)$$

where $U(\cdot)$ is the decision maker's *utility function*: a monotonic increasing function that maps the value of an outcome (e.g., in dollars) to the decision maker's utility for that outcome. Similarly, let $EU(\phi, 0)$ denote the expected utility of the decision maker if he acts immediately, and let $CE(\phi, 0)$ denote the certain equivalent of this situation. Thus, in the myopic procedure, a decision maker should observe the piece of evidence $E$ for which the quantity

$$CE(E, C_E) - CE(\phi, 0) \qquad (10)$$

is maximum, provided it is greater than 0.

In this paper, to simplify the discussion, we assume that the delta property holds.[2] The *delta property* states that an increase in value of all outcomes in a lottery by an amount $\triangle$ increases the certain equivalent of that lottery by $\triangle$ (Howard, 1967). Under this assumption, we obtain

$$CE(E, C_E) = CE(E, 0) - C_E \qquad (11)$$

where $CE(E, 0)$ is the certain equivalent of observing $E$ *at no cost*. Therefore, we have

$$CE(E, C_E) - CE(\phi, 0) = VI(E) - C_E \qquad (12)$$

where

$$VI(E) = CE(E, 0) - CE(\phi, 0) \qquad (13)$$

is the *value of information* of observing $E$.[3] The quantity $VI(E)$ represents the largest amount that the decision maker would be willing to pay to observe $E$. When we compare Expression 10 with Equation 12, we see that, in the myopic procedure, a decision maker should observe the piece of evidence $E$ (if any) for which the quantity

$$VI(E) - C_E \equiv NVI(E) \qquad (14)$$

is maximum and positive. We call $NVI(E)$ the *net value of information* of observing $E$.

The decision maker usually specifies directly the cost of observing evidence. In contrast, we can compute $VI(E)$ from the decision maker's utilities and probabilities. Specifically, from Equations 9 and 13, we have

$$VI(E) = U^{-1}(EU(E, 0)) - U^{-1}(EU(\phi, 0))$$

To simplify notation, we use the abbreviations

$$EU(E, 0) \equiv EU(E) \quad \text{and} \quad EU(\phi, 0) \equiv EU(\phi)$$

Thus, we obtain

$$VI(E) = U^{-1}(EU(E)) - U^{-1}(EU(\phi)) \qquad (15)$$

The computation of $EU(\phi)$ is straightforward. We have

$$EU(\phi) = \begin{cases} p(H)U(H, \neg D) + p(\neg H)U(\neg H, \neg D), \\ \qquad\qquad\qquad\qquad p(H) \leq p^* \\ \\ p(H)U(H, D) + p(\neg H)U(\neg H, D), \\ \qquad\qquad\qquad\qquad p(H) > p^* \end{cases} \qquad (16)$$

by definition of $p^*$.

To compute $EU(E)$, let us assume that $E$ is defined such that the observation of $E$ increases the probability of $H$. If $p(H|E) > p^*$ and $p(H|\neg E) > p^*$, then $VI(E) = 0$, because the decision maker will not change his decision if he observes $E$. Similarly, if $p(H|E) < p^*$ and $p(H|\neg E) < p^*$, then $VI(E) = 0$. Thus, we need only to consider the case where $p(H|E) > p^*$ and $p(H|\neg E) < p^*$. Let us consider separately the cases $H$ and $\neg H$. We have

$$\begin{aligned}EU(E|H) = \\ p(E|H)U(H,D) + p(\neg E|H)U(H, \neg D)\end{aligned} \qquad (17)$$

and

$$\begin{aligned}EU(E|\neg H) = \\ p(E|\neg H)U(\neg H, D) + p(\neg E|\neg H)U(\neg H, \neg D)\end{aligned} \qquad (18)$$

where $EU(E|H)$ and $EU(E|\neg H)$ are the expected utilities of observing $E$, given $H$ and $\neg H$, respectively. To obtain the expected utility of observing $E$, we average these two quantities

$$EU(E) = p(H)EU(E|H) + p(\neg H)EU(E|\neg H) \qquad (19)$$

To compute $VI(E)$, we combine Equations 15, 16, and 19.

---

[2] The primary result of this research—that we can use the central-limit theorem to make tractable an approximate nonmyopic analysis—is unaffected by this assumption.

[3] Other names for $VI(E)$ include the value of perfect information of $E$ and the value of clairvoyance on $E$.

## 4 NONMYOPIC ANALYSIS

As we mentioned in the previous section, the myopic procedure for identifying cost-effective observations includes the incorrect assumption that the decision maker will act after observing only one piece of evidence. This myopic assumption can affect the diagnostic accuracy of an expert system because information gathering might be halted even though there exists some set of features whose value of information is greater that the cost of its observation. For example, a myopic analysis may indicate that no feature is cost effective for observation, yet the value of information for one or more feature pairs (were they computed) could exceed the cost of their observation.

There has been little investigation of the accuracy of myopic analyses. In one analysis, Kalagnanam and Henrion, 1990, showed that a myopic policy is optimal, when the decision maker's utility function $U(\cdot)$ is linear, and the relationship between hypotheses and evidence is deterministic. In an empirical study, Gorry, 1968, demonstrated that the use of a myopic analysis does not diminish significantly the diagnostic accuracy of an expert system for congenital heart disease.

In a correct identification of cost-effective evidence, we should take into account the fact that the decision maker may observe more than one piece of evidence before acting. This computation must consider all possible ordered sequences of evidence observation, and is, therefore, intractable.

Let us consider, however, the following nonmyopic approximation for identifying features that are cost effective to observe. Again, we assume that the delta property holds. First, under the myopic assumption, we compute the net value of information for each piece of evidence. If there is at least one piece of evidence that has a positive net value of information, then we identify for observation the piece of evidence with the highest net value of information. Otherwise, we arrange the pieces of evidence in descending order of their net values of information. Let us label the pieces of evidence $E_1, E_2, \ldots, E_n$, such that $NVI(E_i) > NVI(E_j)$, if and only if $i > j$.

Next, we compute the net value of information of each subsequence of $E_1, E_2, \ldots, E_n$. That is, for $m = 1, 2, \ldots n$, we compute the difference between the value of information for observing $E_1, E_2, \ldots, E_m$, and the cost of observing this sequence of evidence. If any such net value of information is greater than 0, then we identify $E_1$ as a piece of evidence that is cost effective to observe. Once the decision maker has observed $E_1$, we repeat the entire computation described in this section.

This approach does not consider all possible test sequences, but it does overcome one limitation of the myopic analysis. In particular, the method can identify sets of features that are cost effective for observation, even when the observation of each feature alone is not cost effective.

## 5 VALUE OF INFORMATION FOR A SUBSET OF EVIDENCE

As in the myopic analysis, we assume that the decision maker can specify the cost of observing a set of evidence. In this section, we show how we can compute the value of information for a set of evidence from the decision maker's utilities and probabilities.

As in the previous section, let us suppose that the decision maker has the option to observe a particular subset of evidence $\{E_1, E_2, \ldots, E_m\}$ before acting. There are $2^m$ possible instantiations of the evidence in this set, corresponding to the observation of $E_i$ or $\neg E_i$ for every $i$. Let $\mathcal{E}$ denote an arbitrary instantiation; and let $\mathcal{E}_D$ and $\mathcal{E}_{\neg D}$ denote the set of instantiations $\mathcal{E}$ such that $p(H|\mathcal{E}) > p^*$ and $p(H|\mathcal{E}) \leq p^*$, respectively.

The computation of the value of information for the observation of the set $\{E_1, E_2, \ldots, E_m\}$ parallels the myopic computation. In particular, we have

$$
\begin{aligned}
EU(E_1 \ldots E_m) = \\
p(H)EU(E_1 \ldots E_m|H) + \\
p(\neg H)EU(E_1 \ldots E_m|\neg H)
\end{aligned} \quad (20)
$$

where

$$
\begin{aligned}
EU(E_1 \ldots E_m|H) = \\
\left[\sum_{\mathcal{E} \in \mathcal{E}_D} p(\mathcal{E}|H)\right] U(H, D) + \\
\left[\sum_{\mathcal{E} \in \mathcal{E}_{\neg D}} p(\mathcal{E}|H)\right] U(H, \neg D)
\end{aligned} \quad (21)
$$

and

$$
\begin{aligned}
EU(E_1 \ldots E_m|\neg H) = \\
\left[\sum_{\mathcal{E} \in \mathcal{E}_D} p(\mathcal{E}|\neg H)\right] U(\neg H, D) + \\
\left[\sum_{\mathcal{E} \in \mathcal{E}_{\neg D}} p(\mathcal{E}|\neg H)\right] U(\neg H, \neg D)
\end{aligned} \quad (22)
$$

To obtain $VI(E)$, we combine Equations 15, 16, and 20.

When $m$ is small, we can compute directly the sums in Equations 21 and 22. When $m$ is large, we can compute these sums using an approximation that involves the central limit theorem as follows. First we express the sums in terms of weights of evidence. We have

$$\sum_{\mathcal{E} \in \mathcal{E}_D} p(\mathcal{E}|H) = p(W > W^*|H) \quad (23)$$

$$\sum_{\mathcal{E} \in \mathcal{E}_D} p(\mathcal{E}|\neg H) = p(W > W^*|\neg H) \quad (24)$$

$$\sum_{\mathcal{E} \in \mathcal{E}_{\neg D}} p(\mathcal{E}|H)) = 1 - p(W > W^*|H) \quad (25)$$

$$\sum_{\mathcal{E} \in \mathcal{E}_{\neg D}} p(\mathcal{E}|\neg H)) = 1 - p(W > W^*|\neg H) \quad (26)$$

where $W$ and $W^*$ are defined in Equation 7. The term $p(W > W^*|H)$, for example, is the probability that the sum of the weight of evidence from the observation of $E_1, E_2, \ldots, E_m$ exceeds $W^*$. That is, $p(W > W^*|H)$ is the probability that the decision maker will take action $D$ after observing the evidence, given that $H$ is true.

Next, let us consider the weight of evidence for one piece of evidence. We have

| $w_i$ | $p(w_i|H)$ | $p(w_i|\neg H)$ |
|---|---|---|
| $\ln \frac{p(E_i|H)}{p(E_i|\neg H)}$ | $p(E_i|H)$ | $p(E_i|\neg H)$ |
| $\ln \frac{p(\neg E_i|H)}{p(\neg E_i|\neg H)}$ | $p(\neg E_i|H)$ | $p(\neg E_i|\neg H)$ |

To simplify notation, we let $p(E_i|H) = \alpha$ and $p(E_i|\neg H) = \beta$. The expectation and variance of $w$, given $H$ and $\neg H$, are then

$$EV(w|H) = \alpha \ln \frac{\alpha}{\beta} + (1-\alpha) \ln \frac{(1-\alpha)}{(1-\beta)} \quad (27)$$

$$Var(w|H) = \alpha(1-\alpha)\ln^2 \frac{\alpha(1-\beta)}{\beta(1-\alpha)} \quad (28)$$

$$EV(w|\neg H) = \beta \ln \frac{\alpha}{\beta} + (1-\beta) \ln \frac{(1-\alpha)}{(1-\beta)} \quad (29)$$

$$Var(w|\neg H) = \beta(1-\beta)\ln^2 \frac{\alpha(1-\beta)}{\beta(1-\alpha)} \quad (30)$$

Now, we take advantage of the additive property of weights of evidence. The central-limit theorem states that the sum of independent random variables approaches a normal distribution when the number of variables becomes large. Furthermore, the expectation and variance of the sum is just the sum of the expectations and variances of the individual random variables, respectively. Because we have assumed that evidence variables are independent, given $H$ or $\neg H$, the expected value of the sum of the weights of evidence for $E_1, E_2, \ldots, E_m$ is

$$EV(W|H) = \sum_{i=1}^{m} EV(w_i|H) \quad (31)$$

The variance of the sum of the weights is

$$Var(W|H) = \sum_{i=1}^{m} Var(w_i|H) \quad (32)$$

Thus, $p(W|H)$, the probability distribution over $W$, is

$$p(W|H) \sim N(\sum_{i=1}^{m} EV(w_i|H), \sum_{i=1}^{m} Var(w_i|H)) \quad (33)$$

The expression for $\neg H$ is similar.

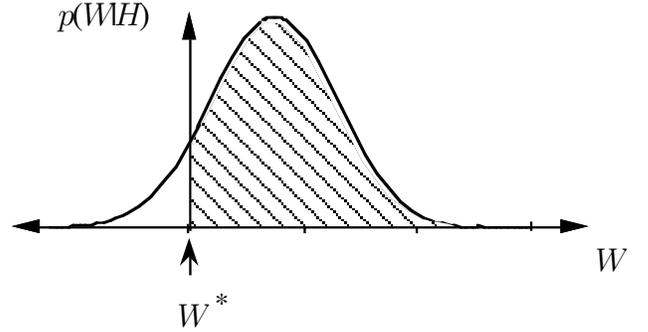

Figure 2: The probability that the total weight of evidence will exceed the threshold weight is the area under the normal curve above the threshold weight $W^*$ (shaded region).

Finally, given the distributions for $H$ and $\neg H$, we evaluate Equations 23 through 26 using an estimate or table of the cumulative normal distribution. We have

$$p(W > W^*|H) = \frac{1}{\sigma\sqrt{2\pi}} \int_{W^*}^{\infty} e^{\frac{-(t-\mu)^2}{2\sigma}} dt \quad (34)$$

where $\mu = EV(W|H)$ and $\sigma = Var(W|H)$. The probability that the weight will exceed $W^*$ corresponds to the shaded area in Figure 2. Again, the expression for $\neg H$ is similar. In this analysis, we assume that no probability ($p(E_i|H)$ or $p(E_i|\neg H)$) is equal to 0 or 1. Thus, all expected values and variances are finite. We relax this assumption in the next section.

## 6 RELAXATION OF THE ASSUMPTIONS

We can relax the assumption that evidence is two-valued with little effort. In particular, we can extend easily the odds-likelihood inference rule, Equation 1, and its logarithmic transform, to include multiple-valued evidential variables. In addition, the computation of means and variances for multiple-valued evidential variables (see Equations 27 through 30) is straightforward.

In addition, we can relax the assumption that no probability is equal to 0 or 1. For example, let us suppose that

$$0 < p(E_j|H) = \alpha < 1$$
$$p(E_j|\neg H) = \beta = 1$$
$$0 < p(E_i|H) < 1, \quad i = 1, 2, \ldots, n \ (i \neq j)$$
$$0 < p(E_i|\neg H) < 1, \quad i = 1, 2, \ldots, n \ (i \neq j)$$

Using Equations 27 through 30, we obtain

$$EV(w_j|H) = +\infty$$

$$Var(w_j|H) = +\infty$$
$$EV(w_j|\neg H) < 0$$
$$Var(w_j|\neg H) = 0$$

Therefore, although the computation of $p(W > W^*|\neg H)$ is straightforward, we cannot compute $p(W > W^*|H)$ as described in the previous section. Instead, we compute $p(W > W^*|H)$, by considering separately the cases $E_j$ and $\neg E_j$. We have

$$\begin{aligned}p(W > W^*|H) &= p(E_j|H)\, p(W > W^*|HE_j) + \\ &\quad p(\neg E_j|H)\, p(W > W^*|H\neg E_j)\end{aligned} \quad (35)$$

If $\neg E_j$ is observed, $W = +\infty$, and $p(W > W^*|H\neg E_j) = 1$. Consequently, Equation 35 becomes

$$\begin{aligned}p(W > W^*|H) &= p(E_j|H)\, p(W > W^*|HE_j) + \\ &\quad p(\neg E_j|H)\end{aligned}$$

We compute $p(W > W^*|HE_j)$ as described in Equations 31 through 34, replacing $EV(w_j|H)$ with $w_j$ in the summation of Equation 31, and $Var(w_j|H)$ with 0 in the summation of Equation 32. The other terms in the summations remain the same, because we have assumed that evidence variables are independent, given $H$ or $\neg H$. This approach generalizes easily to multiple-valued evidence variables and to cases where more than one probability is equal to 0 or 1.

We can extend our analysis to special cases of conditional dependence among evidence variables. For example, Figure 3 shows a schematic of the belief network for Pathfinder. In this model, there are groups of dependent evidence, where each group is conditionally independent of all other groups. We can apply our analysis to this model by using a clustering technique described by Pearl (Pearl, 1988) (pp. 197-204). As in the previous section, suppose we want to compute the value of information for the set of evidence $S = \{E_1, E_2, \ldots, E_m\}$. For each group of dependent features $G^k$, we cluster those variables in the intersection of $S$ and $G^k$ into a single variable. Then, we average out all variables in the belief network that are not in $S$. What remains is a set of clustered variables that are conditionally independent, given $H$ and $\neg H$. We can now apply our analysis—generalized to multiple-valued variables—to this model.

There are special classes of dependent distributions for which the central-limit theorem is valid. We can use this fact to extend our analysis to other cases of dependent evidence. For example, the central-limit theorem applies to distributions that form a Markov chain, provided the transition probabilities in the chain are not correlated (Billingsley, 1968). Thus, we can extend our analysis to belief networks of the form shown in Figure 4. We can generalize the value-of-information analysis even further, if we use the Markov extension in combination with the clustering approach described in the previous paragraph.

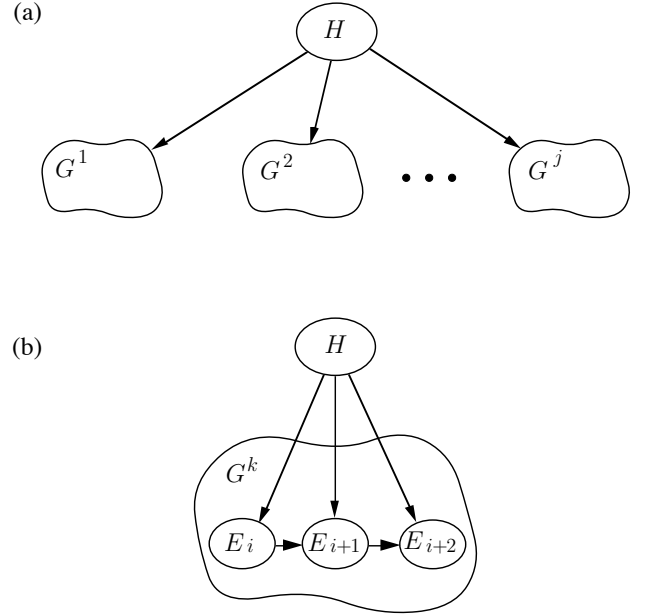

Figure 3: A schematic belief network for Pathfinder. (a) The features in Pathfinder can be arranged into groups of evidence variables $G^1$, $G^2$, ... $G^j$. The variables within each group are dependent, but the groups are conditionally independent, given the disease variable $H$. (b) A detailed view of the evidence variables $E_i$, $E_{i+1}$, and $E_{i+2}$ within group $G^k$.

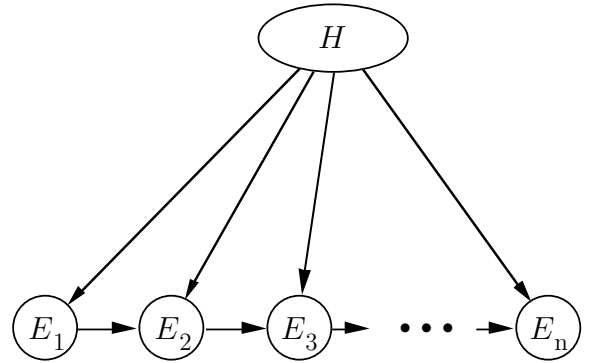

Figure 4: A conditional Markov chain. The evidence variables form a Markov chain conditioned on the variable $H$. We can extend our analysis involving the central-limit theorem to this case.

It is difficult for us to extend the analysis to include multiple-valued hypotheses and decisions. The algebra becomes more complex, because the simple $p^*$ model for action no longer applies. There is, however, the opportunity for applying our technique to more complex problems. In particular, we can abstract a given decision problem into one involving a binary hypothesis and decision variable. For example, we can abstract the problem of determining which of $n$ diseases is present in a patient into one of determining whether the disease is benign or malignant. In doing so, we ignore details of the decision maker's preferences, and we introduce dependencies among evidence variables. Nonetheless, the benefits of a nonmyopic analysis may outweigh these drawbacks in some domains.

## 7  SUMMARY AND CONCLUSIONS

We presented work on the use of the central-limit theorem to compute the value of information for sets of tests. Our technique provides a nonmyopic, yet tractable alternative to traditional myopic analyses for determining the next best piece of evidence to observe. Our approach is limited to information-acquisition decisions for problems involving (1) specific classes of dependencies among evidence variables, and (2) binary hypothesis and action variables. Additional research, however, may help to relax these restrictions. For now, we pose the nonmyopic methodology as a new special-case tool for identifying cost-effective observations. We hope to see empirical comparisons of the relative accuracy of the nonmyopic analysis with that of traditional myopic analyses. We expect that the results of such evaluations will be sensitive to the details of the application areas.

### Acknowledgments

This work was supported by the National Cancer Institute under Grant RO1CA51729-01A1, and by the Agency for Health Care Policy and Research under Grant T2HS00028.